\documentclass{article}

\PassOptionsToPackage{numbers,compress}{natbib}
\usepackage[preprint]{neurips_2026}

\usepackage[utf8]{inputenc}
\usepackage[T1]{fontenc}
\usepackage{hyperref}
\usepackage{url}
\usepackage{booktabs}
\usepackage{amsfonts}
\usepackage{amsmath}
\usepackage{amssymb}
\usepackage{nicefrac}
\usepackage{microtype}
\usepackage[table]{xcolor}
\usepackage{graphicx}
\graphicspath{{figures/}}
\usepackage{enumitem}
\usepackage{pifont}  
\usepackage{wrapfig}  
\usepackage{float}  
\usepackage{algorithm}
\usepackage{algorithmic}
\usepackage{array}  

\newcommand{\Nkeep}{N_{\mathrm{keep}}}

\title{\makebox[\textwidth][c]{\parbox{\dimexpr\textwidth+3cm\relax}{\centering
Beyond Token Importance: Preserving Spatial Scaffolds\\
for Efficient Vision-Language-Action Inference}}}

\author{%
  \makebox[\textwidth][c]{%
    \begin{tabular}{c}
      Jiayu Chen\textsuperscript{1} \quad Shuyong Gao\textsuperscript{2} \quad Jingkai Jia\textsuperscript{1} \quad Xiaosheng Bu\textsuperscript{1} \quad Jiyuan Fu\textsuperscript{1} \\
      Lingyi Hong\textsuperscript{1} \quad Kaixun Jiang\textsuperscript{1} \quad Yipan Xu\textsuperscript{1} \quad Wenqiang Zhang\textsuperscript{1,\ensuremath{\dagger}}
    \end{tabular}%
  }\\[2em]
  \textsuperscript{1}Fudan University \quad \textsuperscript{2}The Hong Kong Polytechnic University \\
  \textsuperscript{\ensuremath{\dagger}} Corresponding author.
}

\begin{document}

\maketitle

\begin{abstract}
Existing VLA pruning strategies primarily select individual visual tokens according to task-level semantic relevance, while overlooking the spatial information required for robotic manipulation. To examine this limitation, we construct a simple \emph{Stride} baseline that uniformly samples tokens along the flattened one-dimensional visual sequence, representing a purely geometric pruning strategy. Surprisingly, Stride outperforms semantic pruning and random pruning at certain pruning ratios, but collapses when the token budget is only slightly reduced. We characterize this phenomenon through the \emph{spatial coverage radius}, defined as the largest spatial blind spot induced by the retained token set after pruning. Our analysis reveals a strong correlation between the spatial structure of retained tokens and task success, suggesting that reliable VLA pruning requires preserving not only task-relevant tokens but also the spatial scaffold of the scene. Motivated by this diagnosis, we propose \emph{GeoScaffold}, a training-free visual token pruning method that partitions each image into spatial regions, allocates inter-region token budgets using task-relevance weights, and selects intra-region scaffold tokens via farthest point sampling to reduce the local coverage radius. On $\pi_{0.5}$ and LIBERO, GeoScaffold retains only $20\%$ of visual tokens while preserving a $93.2\%$ average success rate, and achieves a $1.78\times$ prefill speedup over the unpruned baseline.
\end{abstract}

\section{Introduction}
\label{sec:intro}

Vision-language-action (VLA) models have demonstrated strong capabilities in robotic manipulation by mapping visual observations and language instructions directly to actions~\cite{rt1,rt2,open_x,openvla,octo,rdt1b,gr2,pi0}. In common manipulation scenarios, these models rely on multi-view visual inputs to infer the state of the environment. Encoding multiple images into visual tokens, however, leads to a long multimodal prefix and substantial inference cost. To improve efficiency, recent methods prune, cache, or reuse visual tokens according to task relevance, temporal redundancy, or action-stage dynamics~\cite{vla_cache,adp,efficientvla}. These methods can identify whether individual tokens are important or redundant, but they do not explicitly preserve the spatial structure of the visual scene.

This limitation is particularly important for VLA inference, because robotic manipulation requires not only recognizing task-relevant objects but also localizing them within the scene~\cite{spatialvla,tracevla,3dsvla,geovla,dreamvla,roboground,spatialpolicy}. The outputs of a VLA policy are embodied physical actions, including gripper poses, contact points, and reference frames grounded in scene geometry~\cite{rt2,pi0,spatialvla,tracevla}. Therefore, the retained visual tokens must preserve sufficient spatial information about the environment to support accurate action generation. This motivates a central question: can VLA pruning remain reliable if it preserves individually important tokens but ignores their spatial arrangement?

\begin{figure}[t]
\centering
\makebox[\textwidth][c]{\includegraphics[width=\dimexpr\textwidth+1.5cm\relax]{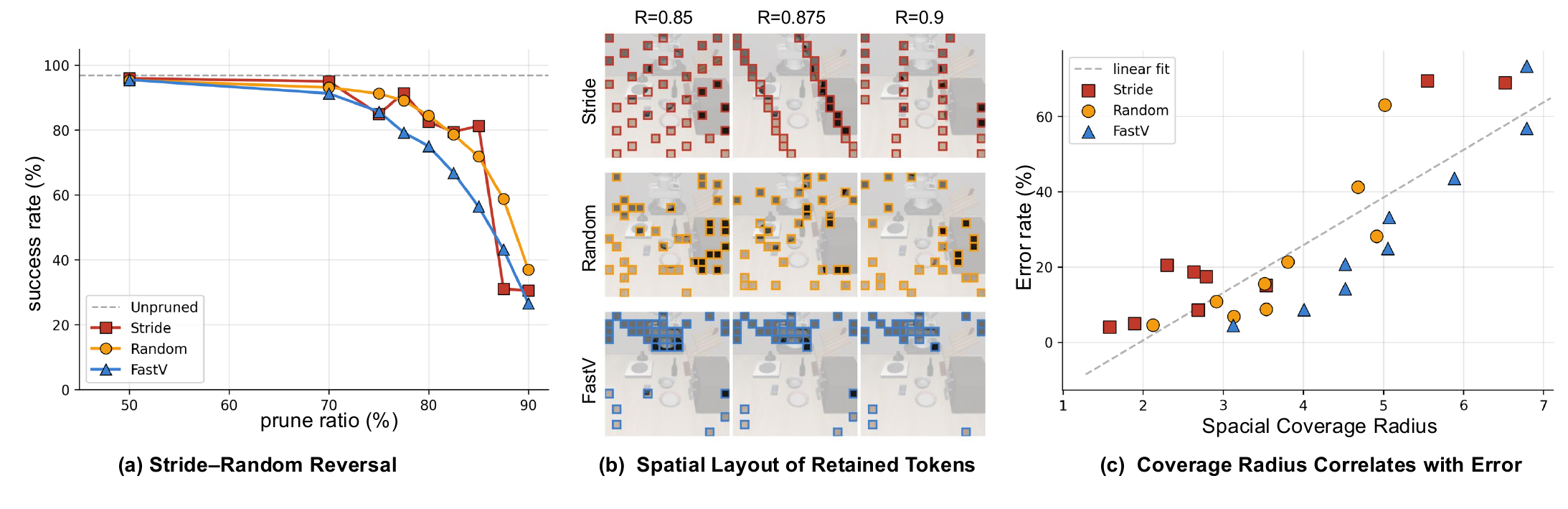}}
\caption{\textbf{(a)} Structured stride pruning can outperform random and attention-based pruning at moderate rates, yet fails abruptly at a nearby pruning rate.
\textbf{(b)} Different pruning policies induce distinct spatial layouts of retained tokens, exposing large uncovered regions under aggressive pruning.
\textbf{(c)} The coverage radius, measuring the largest spatial blind spot, shows a strong positive correlation with pruning-induced error across pruning policies and rates.}
\label{fig:motivation}
\vspace{-3em}
\end{figure}

To examine this question, we construct a minimal \emph{Stride} baseline. It uniformly samples retained tokens along the flattened one-dimensional visual sequence and uses no semantic information; it only reflects the geometric arrangement induced by token positions. Under the conventional view, semantic relevance pruning should consistently outperform such a purely geometric strategy. Surprisingly, this is not what we observe. As shown in Figure~\ref{fig:motivation}(a), Stride substantially outperforms both semantic relevance pruning and random pruning at an $85\%$ pruning ratio, but suffers a sharp performance collapse at a nearby pruning ratio of $87.5\%$. We refer to this empirical phenomenon as the \emph{Stride--Random Reversal}. It suggests that the failure of VLA pruning cannot be explained solely by which tokens are individually important: retained visual tokens must not only be semantically relevant, but also preserve the spatial scaffold of the scene as a set.

To quantify this spatial factor, we introduce the \emph{spatial coverage radius}, defined as the maximum distance from any position on the image grid to its nearest retained token after pruning. This metric characterizes the largest spatial blind spot left by the retained token set. As shown in Figure~\ref{fig:motivation}(b,c), the coverage radius is strongly correlated with pruning-induced error. This diagnosis leads to a simple principle: VLA visual token pruning should preserve both semantic information and geometric coverage. Semantic relevance is better suited for deciding which regions deserve more budget, whereas geometric structure is better suited for deciding which specific tokens should be retained within each region.

Based on this principle, we propose \textbf{GeoScaffold}, a training-free visual token pruning method for VLA inference. GeoScaffold partitions the visual grid of each image into spatial regions. It first estimates region-level semantic weights from shallow layer attention maps and allocates token budgets under a spatial floor constraint that keeps at least one token in each region. Then, within each region, GeoScaffold selects scaffold tokens using center-seeded farthest point sampling, directly reducing the local coverage radius.

Experiments show that GeoScaffold substantially improves over existing semantic relevance pruning methods under aggressive pruning. On $\pi_{0.5}$ and LIBERO, at a pruning ratio of $90\%$, GeoScaffold achieves an average success rate of $75.1\%$ across the four task suites, exceeding the semantic pruning baseline FastV by $48.5$ percentage points, while delivering a $1.96\times$ prefill speedup. The same principle also transfers to OpenVLA-OFT: at a pruning ratio of $87.5\%$, GeoScaffold achieves an average success rate of $79.9\%$, compared with $72.6\%$ for FastV and $51.8\%$ for VLA-Cache. These results suggest that preserving spatial scaffolds is not specific to $\pi_{0.5}$, but is a general requirement for reliable high-ratio VLA pruning.

Our contributions are threefold:
\begin{itemize}[leftmargin=1.5em,topsep=2pt]
  \item We identify the \emph{Stride--Random Reversal}, a counterintuitive phenomenon showing that VLA pruning cannot be understood solely through semantic relevance. The spatial layout of retained tokens becomes a critical variable under aggressive pruning. To quantify this factor, we introduce the \emph{spatial coverage radius}, which measures the largest spatial blind spot induced by pruning and accounts for the observed reversal.
  \item Building on this diagnosis, we propose \emph{GeoScaffold}, a training-free visual token pruning method for efficient VLA inference. GeoScaffold decomposes pruning into semantic-driven inter-region budget allocation and geometry-driven intra-region token selection.
  \item Extensive experiments show that GeoScaffold achieves both higher task success and practical inference acceleration under high pruning ratios. On $\pi_{0.5}$, GeoScaffold retains a $93.2\%$ success rate with a $1.78\times$ prefill speedup at an $80\%$ pruning ratio. We further validate its generality on OpenVLA-OFT.
\end{itemize}

\section{Background}
\label{sec:background}

\subsection{Related Work}
\label{sec:related_work}

\paragraph{Spatially grounded VLAs.}
Vision-language-action (VLA) models extend vision-language priors to embodied control by predicting robot actions from visual observations and language instructions. Representative systems such as RT-2~\cite{rt2}, OpenVLA~\cite{openvla}, OpenVLA-OFT~\cite{openvla_oft}, and $\pi_0$~\cite{pi0} show that large-scale visual-language pretraining can support semantic generalization and end-to-end robot manipulation. However, manipulation requires more than semantic recognition: action prediction depends on spatially grounded perception, including object locations, contact regions, relative geometry, free space, and motion trajectories. Recent spatially aware VLAs make this requirement explicit. SpatialVLA~\cite{spatialvla} injects 3D spatial representations through ego-centric positional encoding and adaptive action grids, while DreamVLA~\cite{dreamvla} models dynamic, spatial, and semantic world knowledge for policy learning. These works motivate our study of whether efficient VLA inference can preserve the sparse spatial structure required for manipulation.

\paragraph{Efficient VLA inference.}
Existing efficient VLA methods reduce inference cost through training-time modification, architecture design, caching, or pruning. DeeR-VLA~\cite{deer_vla} introduces dynamic early exiting to reduce the number of activated MLLM layers. RoboMamba~\cite{robomamba} explores efficient Mamba-based VLA architectures, while Fast-in-Slow~\cite{fast_in_slow} couples slow VLM-based reasoning with fast action execution. In contrast, training-free methods aim to accelerate pretrained VLAs without updating model parameters. VLA-Cache~\cite{vla_cache} exploits temporal continuity by reusing cached key-value representations across adjacent frames. EfficientVLA~\cite{efficientvla} combines layer pruning, task-aware visual token selection, and action-head feature caching. ADP~\cite{adp} further performs action-aware dynamic pruning by adapting the token retention ratio according to recent action trajectories and manipulation stages.

Our work follows the training-free setting but revisits the token-selection principle. Existing VLA pruning methods largely retain tokens by attention, text relevance, task relevance, or action-conditioned saliency, without explicitly preserving their spatial distribution. In manipulation, such importance-only pruning may discard low-saliency yet spatially necessary tokens, weakening object localization, contact reasoning, and trajectory prediction. GeoScaffold therefore combines semantic inter-region allocation with geometric intra-region selection to preserve both task relevance and spatial coverage without additional training.

\subsection{Preliminaries}
\label{sec:prelim}

\paragraph{Dual-system VLA pipeline.}
Many recent vision-language-action (VLA) models~\cite{pi0, pi05, gr00t} follow a dual-system architecture that separates perception-language understanding from continuous action generation. At each control step \(t\), the policy receives a multimodal observation
\begin{equation}
  \mathbf{o}_t =
  \bigl\{\mathbf{I}^1_t,\ldots,\mathbf{I}^n_t,\boldsymbol{\ell}_t,\mathbf{q}_t\bigr\},
\end{equation}
where \(\mathbf{I}^i_t\) denotes the \(i\)-th RGB view, \(\boldsymbol{\ell}_t\) is the language instruction, and \(\mathbf{q}_t\) is the proprioceptive state. Each image is encoded into \(T\) visual tokens arranged on a \(\sqrt{T}\times\sqrt{T}\) patch grid. The model predicts a continuous action chunk
\begin{equation}
  \mathbf{A}_t =
  [\mathbf{a}_t,\mathbf{a}_{t+1},\ldots,\mathbf{a}_{t+H-1}],
\end{equation}
with horizon \(H\).

Conceptually, inference consists of two coupled stages. First, a vision-language backbone \(\mathcal{F}\) encodes the observation into a reusable multimodal prefix:
\begin{equation}
  \mathbf{Z} = \mathcal{F}(\mathbf{o}_t).
\end{equation}
Then, an action expert \(g_\phi\) generates the action chunk by iteratively refining action tokens conditioned on the prefix:
\begin{equation}
  \mathbf{A}^{(s+1)}_t =
  g_\phi\bigl(\mathbf{A}^{(s)}_t,\mathbf{Z}\bigr),
  \qquad s=0,\ldots,T_d-1 .
\end{equation}
Here \(T_d\) denotes the number of action-generation iterations. Since the prefix is computed once but repeatedly attended to during action generation, and visual tokens usually dominate the prefix length, reducing redundant visual tokens provides a direct path to accelerating this architecture class without modifying model parameters.

\paragraph{Visual token pruning.}
Following prior work on visual token pruning in LVLMs and efficient VLA inference~\cite{fastv,sparsevlm,atp_llava,efficientvla}, we formalize visual token pruning as budgeted selection within the visual prefix. Let
\(\mathbf{V}=\{\mathbf{V}^i\}_{i=1}^{n}\) denote the visual tokens from \(n\) camera views, where
\(\mathbf{V}^i=\{\mathbf{v}^i_j\}_{j=1}^{T}\). At an anchor layer \(K\), a pruning policy selects a subset \(\mathbf{S}\subseteq\mathbf{V}\) with a per-view budget
\begin{equation}
  \Nkeep=\max\bigl(1,\lfloor T(1-R)\rfloor\bigr),
  \qquad R\in[0,1),
\end{equation}
such that \(|\mathbf{S}|=n\Nkeep\). The selected tokens replace the original visual tokens in all subsequent layers, while non-visual tokens are kept unchanged.

We focus on \emph{physical pruning}, where discarded tokens are removed from the sequence rather than attention-masked, allowing later transformer layers to operate on a shorter prefix. Ideally, the retained subset should preserve the action chunk predicted by the full model:
\begin{equation}
  \mathbf{S}^{*}
  =
  \arg\min_{\substack{\mathbf{S}\subseteq\mathbf{V}\\|\mathbf{S}|=n\Nkeep}}
  \mathcal{D}\bigl(\mathbf{A}_t,\mathbf{A}_{t,\mathbf{S}}\bigr),
\end{equation}
where \(\mathbf{A}_{t,\mathbf{S}}\) is the prediction after replacing \(\mathbf{V}\) with \(\mathbf{S}\) at layer \(K\), and \(\mathcal{D}\) measures the deviation from the full-model action chunk.
\section{Methodology}
\label{sec:diagnosing}

\subsection{Stride--Random Reversals Reveal Spatial Aliasing}
\label{sec:stride_paradox}

As shown in Figure~\ref{fig:motivation}(a), a one-dimensional sampling strategy that uses no visual semantic information can outperform semantic pruning methods at certain pruning ratios, yet collapse sharply when the token budget changes only slightly. We refer to this empirical pattern as the \emph{Stride--Random reversal}. This observation motivates us to reexamine a basic assumption in VLA visual token pruning: is the choice of retained visual tokens primarily determined by their semantic relevance to the image or task?

To isolate the effect of semantic cues from spatial layout, we introduce a minimal baseline, \emph{Stride}, which serves as a purely geometric selection rule. Given the flattened sequence of visual tokens extracted from an image, Stride retains a fixed number of tokens at uniform intervals along the one-dimensional sequence. Under the common task-relevance view of VLA pruning, such a content-agnostic rule should be consistently weaker than attention-based selection. However, the results of $\pi_{0.5}$ on LIBERO suggest otherwise.

First, a purely geometric strategy can outperform pruning methods that explicitly use semantic information. At a pruning ratio of $85\%$, Stride achieves an average success rate of $81.3\%$ across the four LIBERO task suites, outperforming both Random selection at $71.8\%$ and semantic pruning at $56.4\%$. This advantage is not confined to a single task suite: at the same pruning ratio, Stride outperforms both baselines on LIBERO-Spatial, Object, Goal, and Long, as shown in Figure~\ref{fig:figure1a_grid} in Appendix~\ref{app:libero}. Since Stride has no access to which tokens are task-relevant, this result cannot be simply attributed to better importance estimation.

Second, this advantage is highly sensitive to small changes in the token budget. When the pruning ratio increases from $85\%$ to $87.5\%$, the per-view budget decreases from $38$ to $32$ retained tokens, yet the average success rate of Stride drops from $81.3\%$ to $31.0\%$, a decrease of $50.3$ percentage points. In contrast, Random selection and semantic pruning degrade more smoothly over the same interval. Therefore, this failure is not merely a consequence of retaining fewer tokens; rather, it appears to be closely related to the spatial layout induced by the retained tokens on the image grid.

Third, stronger pruning does not always lead to worse performance. The Stride curve is not monotonic: when the pruning ratio increases from $75\%$ to $77.5\%$, the average success rate rises from $84.9\%$ to $91.4\%$, despite retaining fewer tokens. A similar local recovery also appears from $82.5\%$ to $85\%$. This suggests that VLA pruning performance is not governed solely by the number of retained tokens. Small changes in the token budget may induce substantially different spatial coverage patterns, leading to markedly different downstream behavior.

The Stride--Random reversal suggests that pruning performance depends not only on which individual tokens are selected, but also on how the retained tokens are spatially arranged as a set. In other words, a pruning strategy must preserve sufficient spatial support to represent the scene geometry.

\subsection{Quantifying Spatial Scaffold Collapse via Coverage Radius}
\label{sec:coverage_radius}

Motivated by the Stride--Random reversal, we next ask whether the spatial structure of retained tokens can be characterized by a simple geometric quantity. In this section, we first define \emph{spatial coverage radius}, which measures the largest spatial blind spot left by pruning. We then examine its relationship with pruning-induced error, and finally discuss how this geometric metric turns the empirical reversal into an actionable design principle.

\paragraph{Spatial coverage radius.}
For an image, we view the retained visual tokens as a set of points on the two-dimensional patch grid, where each point corresponds to the center of a retained image patch. The \emph{spatial coverage radius} is defined as the maximum distance from any location on the image grid to its nearest retained token:
\begin{equation}
  \rho(\mathcal{S})
  =
  \max_{\mathbf{x}\in\mathcal{G}}
  \min_{\mathbf{s}\in\mathcal{S}}
  d(\mathbf{x}, \mathbf{s}),
\end{equation}
where $\mathcal{G}$ denotes the full image patch grid, $\mathcal{S}\subseteq\mathcal{G}$ denotes the retained token set, and $d(\cdot,\cdot)$ is the Euclidean distance on the grid.

Intuitively, $\rho(\mathcal{S})$ measures the size of the largest spatial blind spot induced by pruning. A smaller radius indicates that the retained tokens provide more uniform support over the image space. A larger radius, in contrast, suggests that some regions lack direct visual evidence, forcing the action decoder to infer their state from more distant tokens.

\paragraph{Coverage radius correlates with pruning failure.}
We compute the spatial coverage radius for $\pi_{0.5}$ on LIBERO across different pruning strategies and pruning ratios, and compare it with the task failure rate. The pruning strategies include Stride, Random, and Semantic-Relevance Pruning, whose retained-token layouts are illustrated in Figure~\ref{fig:motivation}(b). As shown in Figure~\ref{fig:motivation}(c), coverage radius is strongly correlated with pruning failure: larger spatial blind spots generally correspond to higher error rates. Across $27$ method--ratio configurations, the Pearson correlation between coverage radius and failure rate is $0.85$, and the Spearman correlation is $0.84$. This indicates that, even without using image semantics, the spatial layout of retained tokens explains a substantial portion of the performance variation after pruning.

We further compare different strategies under the same pruning ratio. In this setting, all methods retain the same number of tokens, so the performance difference mainly comes from how these tokens are spatially arranged. Averaging the correlation over $9$ pruning ratios yields a correlation coefficient of $0.93$, further suggesting that spatial coverage is a strong indicator of pruning quality when the token budget is fixed.

\paragraph{Explaining the Stride--Random reversal.}
Coverage radius explains not only the overall trend, but also the key reversal phenomenon observed in Section~\ref{sec:stride_paradox}. For each pruning ratio, we compare pairs of pruning strategies and test a simple rule: the method with the smaller spatial blind spot should achieve the higher success rate. At the three pruning ratios where the Stride--Random reversal is most pronounced, coverage radius gives the correct ordering. At an $85\%$ pruning ratio, Stride has a smaller coverage radius than Random and achieves a higher success rate. When the pruning ratio increases to $87.5\%$, however, the one-dimensional stride pattern aliases poorly with the two-dimensional grid, producing larger uncovered regions; correspondingly, Stride performs substantially worse than Random. The same direction of explanation also holds at the $90\%$ pruning ratio.

These results suggest that the fluctuation of Stride is not accidental. Rather, it is driven by the quality of spatial coverage induced by the retained token set. Overall, among $27$ pairwise comparisons, coverage radius correctly ranks $22$ pairs. The remaining $5$ mismatches mainly occur in low-pruning regimes, where success rates are close to saturation and the discriminative signal of spatial layout becomes weaker.

\paragraph{From metric to design principle.}
Spatial coverage radius is not intended to be the only factor determining VLA performance. Task relevance, camera-view differences, and the current stage of manipulation can all affect the final success rate. Nevertheless, coverage radius reveals a missing constraint in existing relevance-based pruning methods: before visual tokens are selected as individually important, they must first preserve sufficient spatial support as a set. Therefore, effective VLA pruning should preserve both semantic relevance and geometric coverage.

\begin{figure}[t]
\centering
\includegraphics[width=\linewidth,height=0.22\textheight,keepaspectratio]{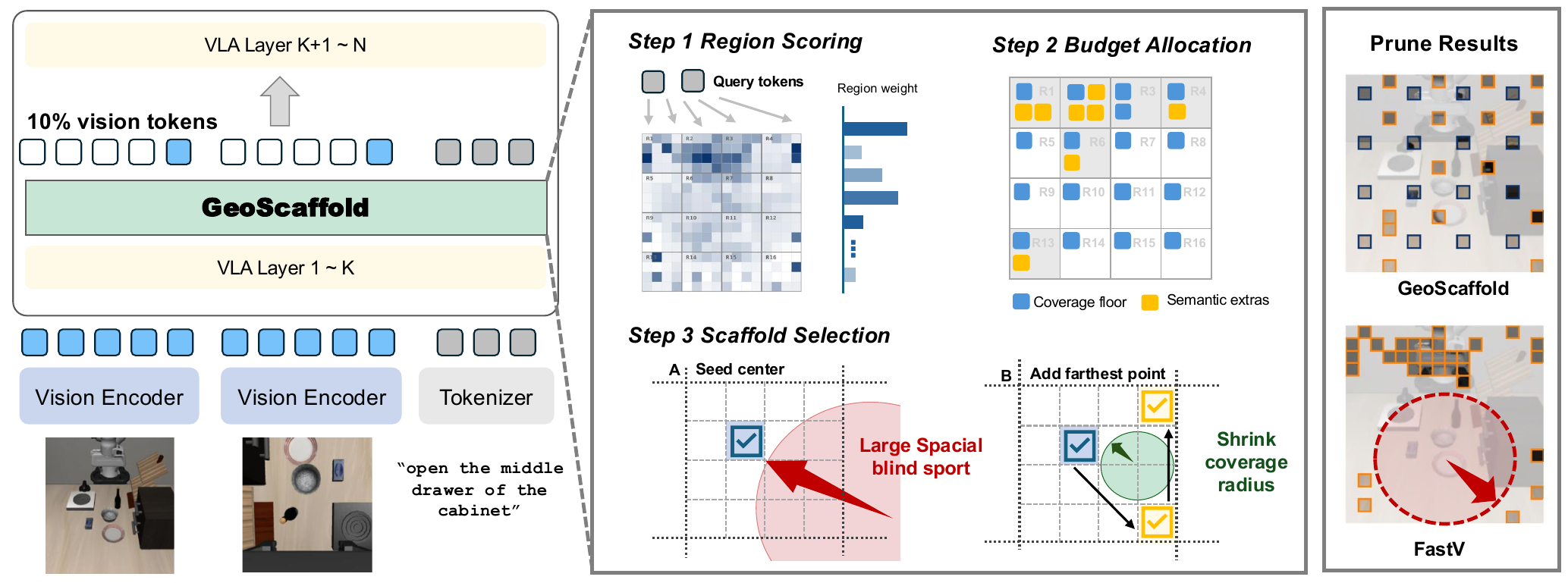}
\caption{Overview of GeoScaffold. Three steps implement the two-level decomposition: \textbf{Step 1: Region Scoring}, aggregate token attention into region weights; \textbf{Step 2: Budget Allocation}, assign budgets with a one-token spatial floor per region; \textbf{Step 3: Scaffold Selection}, use center-seeded farthest point sampling to minimize local coverage radius.}
\label{fig:pipeline}
\vspace{-1em}
\end{figure}

\subsection{GeoScaffold}
\label{sec:method}

We instantiate the above principle as \emph{GeoScaffold}, a simple training-free pruning method for vision-language-action models that preserves both semantic relevance and geometric coverage through a two-level decomposition. GeoScaffold only uses the shallow-layer attention maps already computed by the model, following the FastV setting~\cite{fastv} by default using the second layer. It introduces no trainable parameters and does not modify the model architecture. As illustrated in Figure~\ref{fig:pipeline}, GeoScaffold consists of three steps. By default, we partition the visual grid of each image into non-overlapping regions. For a $16\times16$ grid with $256$ visual tokens, we divide the grid into $N\times N$ regions, each containing the same number of tokens, with $N=4$ by default.

Formally, let $\mathcal{G}$ denote the full visual-token grid and let $\{\mathcal{R}_r\}_{r=1}^{M}$ be a partition of $\mathcal{G}$ into $M=N^2$ disjoint regions:
\begin{equation}
  \mathcal{G} = \bigcup_{r=1}^{M} \mathcal{R}_r,
  \qquad
  \mathcal{R}_r \cap \mathcal{R}_{r'} = \emptyset \quad (r\neq r').
\end{equation}
Given a per-view retained-token budget $K$, GeoScaffold selects a subset
$\mathcal{T}\subseteq\mathcal{G}$ with $|\mathcal{T}|=K$.

\paragraph{Step 1: From token importance to region-level semantic budgets.}
We use the query-to-visual-token attention scores produced at a specified layer of the model and average them across attention heads. Let $A\in\mathbb{R}^{T_q\times T}$ denote the head-averaged query-to-visual-token attention block at the pruning layer, where $T_q$ is the number of query tokens and $T$ is the number of visual tokens. To obtain a more stable estimate of task relevance, GeoScaffold adopts \emph{Semantic Aggregation Attention}, which aggregates attention scores from multiple discriminative text tokens rather than relying on a single query. A text query with higher attention variance tends to concentrate on a small number of salient image patches rather than spreading uniformly over the entire scene, and is therefore more suitable as a discriminative semantic signal. We thus measure the variance of each query's attention over visual tokens and select the top $\rho{=}0.2$ text tokens as the aggregation set $\mathcal{Q}$:
\begin{equation}
  \nu_j = \mathrm{Var}_{i=1}^{T}\!\left(A_{j,i}\right),
  \qquad
  \mathcal{Q} = \operatorname{Top}_{\rho T_q}\!\left(\{\nu_j\}_{j=1}^{T_q}\right).
\end{equation}
The relevance score of each visual token is then computed as the average attention assigned to it by this set, $s_i = \frac{1}{|\mathcal{Q}|}\sum_{j\in\mathcal{Q}} A_{j,i}$.

The key difference from FastV, SparseVLM, and similar importance-based methods is that GeoScaffold does not directly feed these token scores into a token-level top-$k$ selection rule. Instead, it aggregates token scores into region-level signals, $w_r = \sum_{i\in\mathcal{R}_r} s_i$, $r=1,\ldots,M$. In other words, attention is used to estimate how important each region is as a whole. In GeoScaffold, attention is deliberately weakened from a token selector to a region allocator: it no longer answers ``which token is important,'' but rather ``which region is important.''

\paragraph{Step 2: Budget allocation with a spatial floor.}
Given the region weights, GeoScaffold allocates the retained-token budget according to two complementary principles. The first is a \emph{spatial floor}: each region retains at least one token, providing a minimal geometric guarantee. The second is \emph{semantic allocation}: regions with larger weights receive more tokens. Concretely, when $K\ge M$, each region first receives one base token, and the residual budget $K_{\mathrm{rem}}=K-M$ is then distributed in proportion to the region weights:
\begin{equation}
  \tilde{n}_r
  =
  \frac{w_r}{\sum_{r'=1}^{M} w_{r'}}\,K_{\mathrm{rem}},
  \qquad
  n_r = 1 + \left\lfloor \tilde{n}_r \right\rfloor .
\end{equation}
The remaining tokens caused by rounding are assigned to regions with the largest fractional residuals until $\sum_{r=1}^{M} n_r = K$. This spatial floor bounds the worst-case blind-spot radius within each region, serving as an algorithmic expression of the coverage-radius diagnosis in Section~\ref{sec:coverage_radius}.

\paragraph{Step 3: Intra-region scaffold selection by farthest point sampling.}
Within each region, GeoScaffold directly targets the coverage radius identified above as a key predictor of pruning failure. The objective is to select a fixed number of tokens inside a region such that the maximum uncovered distance within that region is as small as possible. Formally, for each region $\mathcal{R}_r$, we aim to solve the discrete $n_r$-center problem:
\begin{equation}
  \mathcal{S}_r^{*}
  =
  \arg\min_{\mathcal{S}\subseteq\mathcal{R}_r,\,|\mathcal{S}|=n_r}
  \rho(\mathcal{S};\mathcal{R}_r),
  \qquad
  \rho(\mathcal{S};\mathcal{R}_r)
  =
  \max_{\mathbf{x}\in\mathcal{R}_r}
  \min_{\mathbf{s}\in\mathcal{S}}
  d(\mathbf{x},\mathbf{s}) .
\end{equation}
We approximate this objective with center-seeded farthest point sampling (FPS). Letting $\mathbf{p}_i$ denote the 2D grid coordinate of token $i\in\mathcal{R}_r$ and $\mathbf{c}_r=\tfrac{1}{|\mathcal{R}_r|}\sum_{i\in\mathcal{R}_r}\mathbf{p}_i$ the region center, FPS proceeds in two stages:
\begin{itemize}[leftmargin=1.5em,topsep=2pt]
  \item \textbf{Center seed:} $q_0 = \arg\min_{i\in\mathcal{R}_r}\|\mathbf{p}_i-\mathbf{c}_r\|_2$, $\mathcal{S}_r=\{q_0\}$;
  \item \textbf{Greedy step:} $q_k = \arg\max_{i\in\mathcal{R}_r\setminus\mathcal{S}_r}\,\min_{j\in\mathcal{S}_r}\|\mathbf{p}_i-\mathbf{p}_j\|_2$, $\mathcal{S}_r\leftarrow\mathcal{S}_r\cup\{q_k\}$, until $|\mathcal{S}_r|=n_r$.
\end{itemize}
The final retained token set is $\mathcal{T} = \bigcup_{r=1}^{M} \mathcal{S}_r$ with $|\mathcal{T}|=K$. Unlike variants that only mask tokens in attention while keeping the sequence length unchanged, GeoScaffold physically removes unselected tokens after the specified layer from the hidden states, attention mask, and position indices. This physical removal is what enables actual acceleration during prefill inference.

\section{Experiments}
\label{sec:experiments}

\paragraph{Experimental setup.}
We use $\pi_{0.5}$~\cite{pi05} as the primary model and further evaluate on OpenVLA-OFT~\cite{openvla_oft} to examine the cross-architecture generality of our method. Both models are evaluated on LIBERO~\cite{libero}, which consists of four task suites: LIBERO-Spatial, Object, Goal, and Long. For each subtask, we run $50$ trials. Success-rate experiments are conducted on NVIDIA A100 GPUs, while end-to-end latency is measured on a single NVIDIA RTX 4090 GPU.

\paragraph{Main results on $\pi_{0.5}$.}
Table~\ref{tab:main_results} reports the performance and inference efficiency of $\pi_{0.5}$ on LIBERO under three pruning rates, $R \in \{0.8, 0.875, 0.9\}$. Since several prior training-free pruning methods~\cite{vla_cache,adp} were designed for the OpenVLA architecture, while this paper focuses on \textbf{dual-system VLAs}, we compare the unpruned model with two semantic relevance pruning baselines, FastV~\cite{fastv} and SparseVLM~\cite{sparsevlm}, as well as our proposed GeoScaffold. We report success rates on the four LIBERO task suites and measure the prefill and end-to-end speedup achieved by GeoScaffold under different pruning rates.

\begin{table}[t]
\caption{Comparison of different VLA acceleration methods on the LIBERO benchmark.}
\label{tab:main_results}
\centering
\small
\setlength{\tabcolsep}{4pt}
\renewcommand{\arraystretch}{1.10}
\begin{tabular*}{\textwidth}{@{\extracolsep{\fill}}lccccc>{\centering\arraybackslash}p{1.6cm}>{\centering\arraybackslash}p{2.2cm}@{}}
\toprule
Method & Spatial & Object & Goal & Long & Avg. & prefill (ms)$\downarrow$ & total (ms)$\downarrow$ \\
\midrule
\rowcolor{gray!15}
\multicolumn{8}{c}{\textit{keep $256/256$ tokens per image, pruning rates = \textbf{\textcolor{blue}{0\%}}}}\\
$\pi_{0.5}$ baseline             & 98.6  & 98.6  & 98.6  & 92.4  & $\mathbf{97.1}$  & 27.68 & 61.03 \\
\midrule
\rowcolor{gray!15}
\multicolumn{8}{c}{\textit{keep $51/256$ tokens per image, pruning rates = \textbf{\textcolor{blue}{80\%}}}}\\
FastV~\cite{fastv}             & 75.4  & 86.6  & 74.0  & 64.2  & 75.0  & 13.51 & 45.52 \\
SparseVLM~\cite{sparsevlm}     & 39.4  & 80.6  & 66.2  & 52.8  & 59.8  & 13.53 & 45.51 \\
\textbf{GeoScaffold (ours)}    & \textbf{97.6} & \textbf{96.6} & \textbf{90.6} & \textbf{88.2} & \textbf{93.2} & 15.58\rlap{\,{\scriptsize\textit{\textbf{\textcolor{blue}{($\times$1.78)}}}}} & 47.56\rlap{\,{\scriptsize\textit{\textbf{\textcolor{blue}{($\times$1.28)}}}}} \\
\midrule
\rowcolor{gray!15}
\multicolumn{8}{c}{\textit{keep $32/256$ tokens per image, pruning rates = \textbf{\textcolor{blue}{87.5\%}}}}\\
FastV~\cite{fastv}             & 38.0  & 62.8  & 44.8  & 27.0  & 43.2  & 12.10 & 43.90 \\
SparseVLM~\cite{sparsevlm}     & 10.2  & 46.4  & 39.8  & 21.4  & 29.5  & 12.12 & 43.96 \\
\textbf{GeoScaffold (ours)}    & \textbf{86.4} & \textbf{95.8} & \textbf{85.2} & \textbf{77.4} & \textbf{86.2} & 14.22\rlap{\,{\scriptsize\textit{\textbf{\textcolor{blue}{($\times$1.95)}}}}} & 46.10\rlap{\,{\scriptsize\textit{\textbf{\textcolor{blue}{($\times$1.32)}}}}} \\
\midrule
\rowcolor{gray!15}
\multicolumn{8}{c}{\textit{keep $25/256$ tokens per image, pruning rates = \textbf{\textcolor{blue}{90\%}}}}\\
FastV~\cite{fastv}             & 15.8  & 41.2  & 35.8  & 13.6  & 26.6  & 12.00 & 43.88 \\
SparseVLM~\cite{sparsevlm}     & 3.6   & 24.4  & 26.4  & 5.6   & 15.0  & 12.03 & 43.85 \\
\textbf{GeoScaffold (ours)}    & \textbf{68.2} & \textbf{90.8} & \textbf{75.6} & \textbf{65.8} & \textbf{75.1} & 14.11\rlap{\,{\scriptsize\textit{\textbf{\textcolor{blue}{($\times$1.96)}}}}} & 45.92\rlap{\,{\scriptsize\textit{\textbf{\textcolor{blue}{($\times$1.33)}}}}} \\
\bottomrule
\end{tabular*}
\end{table}

\paragraph{Semantic relevance pruning collapses under aggressive pruning.}
FastV exhibits a monotonic success-rate drop as the pruning rate increases, decreasing from $75.0\%$ at $R=0.8$ to $26.6\%$ at $R=0.9$, a decline of $48.4$ percentage points. This degradation is most severe on spatially demanding tasks: LIBERO-Spatial drops from $75.4\%$ to $15.8\%$, while LIBERO-Long drops from $64.2\%$ to $13.6\%$. This trend is consistent with our spatial-scaffold perspective and the coverage-radius mechanism discussed in Section~\ref{sec:coverage_radius}. SparseVLM follows the same collapse pattern, suggesting that simply replacing the token evaluator is insufficient to preserve the spatial scaffold under aggressive pruning. Rather, the failure mode is structural: semantic relevance alone does not guarantee spatially faithful token retention.

\paragraph{GeoScaffold remains stable at high pruning rates.}
GeoScaffold maintains strong robustness in the aggressive pruning regime, and its advantage becomes larger as the pruning rate increases. Compared with the strongest existing baseline, FastV, GeoScaffold improves the average success rate by $18.2$ percentage points at $R=0.8$ and by $48.5$ percentage points at $R=0.9$. Even under the most aggressive setting, where only $10\%$ of visual tokens are retained, i.e., $25$ tokens per image, GeoScaffold still achieves a $75.1\%$ average success rate. On LIBERO-Object, GeoScaffold further preserves a $90.8\%$ success rate at $R=0.9$, substantially outperforming semantic relevance pruning baselines.

\paragraph{Inference efficiency.}
\label{sec:exp_latency}
Under aggressive pruning, GeoScaffold also delivers substantial practical speedups. In implementation, the intra-region farthest point sampling step can be optimized through vectorized queries, introducing negligible computation and latency overhead. On a single RTX 4090 GPU, GeoScaffold reduces the prefill latency from $27.68$ ms for the unpruned baseline to $15.58$ ms at $R=0.8$, corresponding to a $1.78\times$ prefill speedup. It also achieves a $1.28\times$ end-to-end speedup, while incurring only a $4\%$ success-rate drop. We refer to Appendix~\ref{app:latency} for the full per-stage latency breakdown and complexity analysis.

\paragraph{Cross-architecture validation.}
\label{sec:oft}
To examine whether the proposed principle generalizes beyond $\pi_{0.5}$, we apply the same pruning strategies to OpenVLA-OFT~\cite{openvla_oft}. As shown in Table~\ref{tab:oft_results} in Appendix~\ref{app:oft}, the resulting performance gap follows a similar structure to that observed on $\pi_{0.5}$. At a pruning rate of $R=0.875$, GeoScaffold achieves a $79.9\%$ average success rate across the four LIBERO suites, outperforming FastV ($72.6\%$) and VLA-Cache~\cite{vla_cache} ($51.8\%$). The advantage is particularly pronounced on LIBERO-Long, where GeoScaffold reaches $66.4\%$, substantially higher than VLA-Cache ($26.2\%$) and FastV ($40.4\%$). These results suggest that preserving spatial scaffolds is not tied to a specific VLA architecture, but provides a general principle for robust visual token pruning in embodied control.

\section{Ablation}
\label{sec:ablation}

\begin{table}[t]
\caption{Ablation studies at $R{=}0.85$ on $\pi_{0.5}$+LIBERO (4-suite avg SR, \%).}
\label{tab:ablation_combined}
\centering
\scriptsize
\setlength{\tabcolsep}{3pt}
\renewcommand{\arraystretch}{1.05}
\begin{minipage}[t]{0.38\textwidth}
\centering
\textit{(a) Single-level baselines.}\\[0.3em]
\begin{tabular}{lcccc}
\toprule
Method & Floor & Geom. & Attn. & SR \\
\midrule
Random              & \ding{55} & \ding{55} & \ding{55} & 71.9 \\
Global-FPS          & \ding{55} & \ding{51} & \ding{55} & 64.2 \\
Stratified-Random   & \ding{51} & \ding{55} & \ding{55} & 75.4 \\
\textbf{GeoScaffold} & \ding{51} & \ding{51} & \ding{51} & \textbf{90.5} \\
\bottomrule
\end{tabular}
\end{minipage}\hfill
\begin{minipage}[t]{0.30\textwidth}
\centering
\textit{(b) Where attention belongs.}\\[0.3em]
{\renewcommand{\arraystretch}{1.30}%
\begin{tabular}{llc}
\toprule
Variant & Intra-region rule & SR \\
\midrule
Region-Semantic       & attention top-$k$    & 72.0 \\
Region-Stride         & 1D linspace          & 81.8 \\
\textbf{GeoScaffold}  & 2D FPS (center)      & \textbf{90.5} \\
\bottomrule
\end{tabular}}
\end{minipage}\hfill
\begin{minipage}[t]{0.28\textwidth}
\centering
\textit{(c) Geometric factorial.}\\[0.3em]
\begin{tabular}{llc}
\toprule
Seed & Selection & SR \\
\midrule
corner & linspace & 81.8 \\
center & linspace & 86.4 ($+4.6$) \\
corner & 2D FPS   & 87.2 ($+5.4$) \\
\textbf{center} & \textbf{2D FPS} & \textbf{90.5} ($+8.7$) \\
\bottomrule
\end{tabular}
\end{minipage}
\end{table}

\paragraph{Can single-level geometric pruning suffice?}
Table~\ref{tab:ablation_combined}(a) compares GeoScaffold with three single-level alternatives at a pruning rate of $R=0.85$: Random, which uniformly samples visual tokens; Global-FPS, which applies farthest point sampling over the entire image grid; and Stratified-Random, which randomly samples tokens within predefined regions. All three alternatives underperform GeoScaffold by at least $15$ percentage points. Notably, Global-FPS is $7.7$ percentage points worse than Random at $R=0.85$, indicating that single-level geometric optimality does not necessarily yield an effective spatial scaffold. Stratified-Random improves over Random by $3.5$ percentage points, suggesting that region partitioning alone provides only moderate gains. The main benefit of GeoScaffold instead comes from the combination of semantic inter-region budget allocation and geometric intra-region farthest point sampling.

\paragraph{Should attention operate at the token level or the region level?}
Table~\ref{tab:ablation_combined}(b) studies where semantic attention should be used. We first consider Region-Semantic, which allocates region-level budgets but still selects tokens within each region by semantic relevance. This strategy achieves a $72.0\%$ success rate. Replacing token-level semantic selection with one-dimensional evenly spaced selection within each region, denoted as Region-Stride, increases the success rate to $81.8\%$. Further upgrading the intra-region selection rule to the two-dimensional center-seeded farthest point sampling used by GeoScaffold improves the success rate to $90.5\%$. These results suggest that attention is more effective when used for inter-region budget allocation, rather than for token-level selection within each region.

\paragraph{What kind of spatial scaffold is needed?}
Table~\ref{tab:ablation_combined}(c) further decomposes the geometric rule in GeoScaffold through a $2\times2$ factorial experiment. We vary two factors: whether the seed location is placed at a corner or at the region center, and whether the selection rule uses one-dimensional stride sampling or two-dimensional farthest point sampling. The two factors contribute approximately additively. Using only center seeding improves performance by $4.6$ percentage points, while using only two-dimensional farthest point sampling improves performance by $5.4$ percentage points. Combining the two yields an $8.7$ percentage-point improvement over Region-Stride at $R=0.85$. Center seeding mainly affects the case where only one token is retained in a region, which is common under high pruning rates. In contrast, two-dimensional farthest point sampling mainly affects regions that retain multiple tokens, which becomes more influential under low-to-moderate pruning rates.

\section{Conclusion}
\label{sec:conclusion}

We revisited visual token pruning for vision-language-action models and found that the dominant semantic-relevance paradigm fails systematically under aggressive pruning. The Stride--Random reversal shows that a content-agnostic baseline can surpass attention-based pruning at certain rates while collapsing under nearby pruning ratios, indicating that downstream success depends not only on which tokens are retained but also on the spatial layout they form as a set. We quantified this factor with the spatial coverage radius and showed that it is the strongest single-variable predictor of pruning-induced error across diverse strategies and pruning rates. Building on this diagnosis, we proposed GeoScaffold, a training-free method that follows a two-level decomposition---attention allocates budget across regions, geometry selects tokens within each region---and combines a per-region spatial floor with center-seeded farthest point sampling. On $\pi_{0.5}$+LIBERO at a $90\%$ pruning rate, GeoScaffold improves the four-suite average success rate by $48.5$ percentage points over FastV while delivering a $1.96\times$ prefill speedup, and the same method transfers to OpenVLA-OFT without modification. We expect this view to extend naturally to video understanding, 3D point cloud subsampling, and other settings where preserving spatial scaffolds is equally load-bearing.


\small
\bibliographystyle{plainnat}
\bibliography{references}
\normalsize

\appendix
\section*{Appendix}
\begin{itemize}
    \item Appendix~\ref{app:libero}: additional per-suite LIBERO results.
    \item Appendix~\ref{app:oft}: cross-architecture results on OpenVLA-OFT.
    \item Appendix~\ref{app:latency}: complexity and latency analysis.
    \item Appendix~\ref{sec:limitations}: limitations and assumptions.
    \item Appendix~\ref{sec:broader_impact}: broader impact discussion.
    \item Appendix~\ref{app:licenses}: assets, licenses, and usage.
\end{itemize}

\section{Additional LIBERO results}
\label{app:libero}

\begin{figure}[H]
\centering
\includegraphics[width=\textwidth]{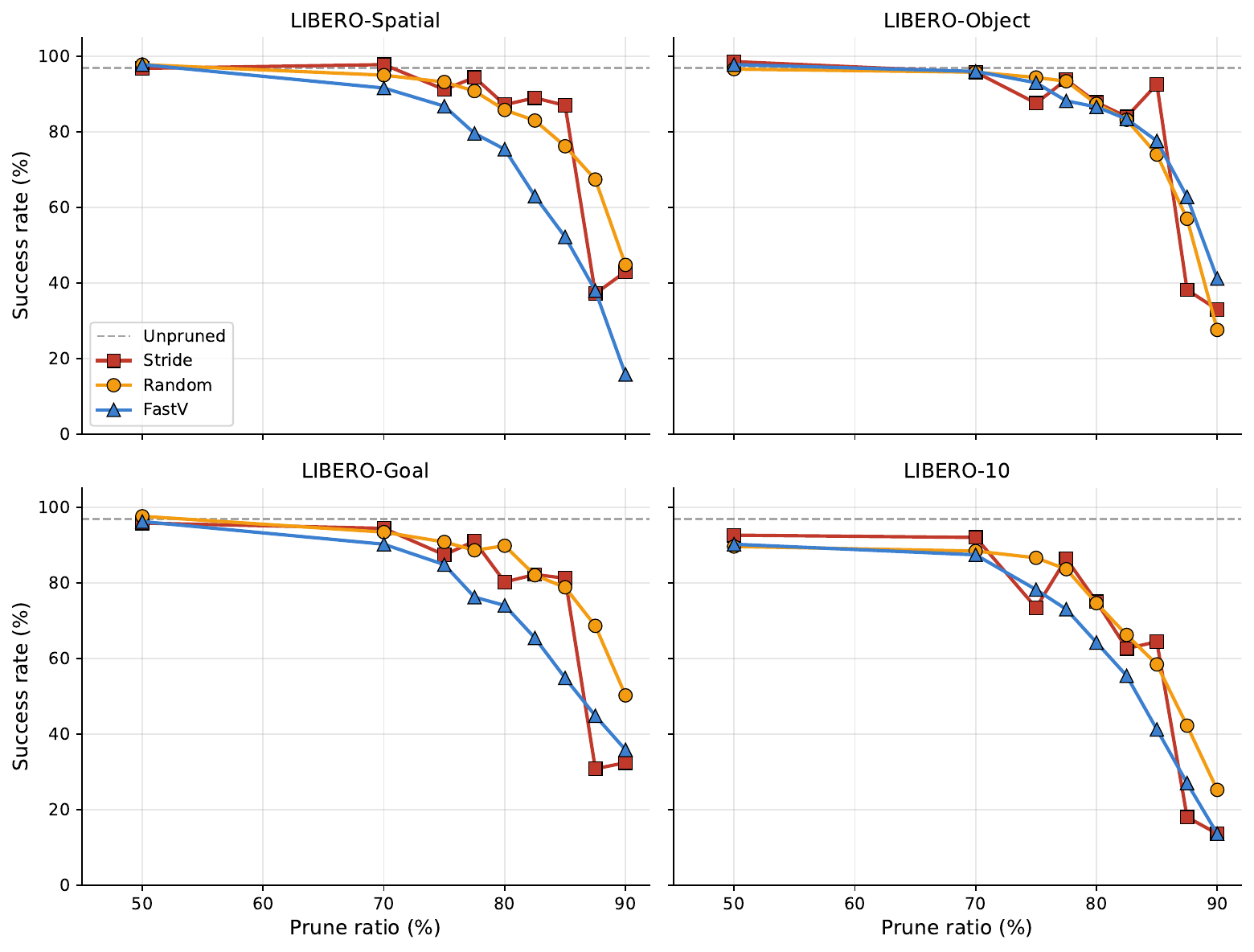}
\caption{Per-suite success rate on $\pi_{0.5}$+LIBERO across pruning ratios for Stride, Random, and semantic pruning. The Stride--Random reversal appears consistently across all four task suites.}
\label{fig:figure1a_grid}
\end{figure}

\section{Cross-architecture results on OpenVLA-OFT}
\label{app:oft}

\begin{table}[H]
\caption{Performance on LIBERO at OpenVLA-OFT~\cite{openvla_oft} across two aggressive pruning rates. Bold = column best within each block.}
\label{tab:oft_results}
\centering
\small
\setlength{\tabcolsep}{6pt}
\renewcommand{\arraystretch}{1.10}
\begin{tabular*}{\textwidth}{@{\extracolsep{\fill}}lccccc@{}}
\toprule
Method & Spatial & Object & Goal & Long & Avg.\\
\midrule
\rowcolor{gray!15}
\multicolumn{6}{c}{\textit{keep $512/512$ tokens per image, pruning rates = \textbf{\textcolor{blue}{0\%}}}}\\
OpenVLA-OFT baseline                       & 95.8 & 98.7 & 96.3 & 90.7 & 95.4 \\
\midrule
\rowcolor{gray!15}
\multicolumn{6}{c}{\textit{keep $128/512$ tokens per image, pruning rates = \textbf{\textcolor{blue}{75\%}}}}\\
VLA-Cache~\cite{vla_cache}                 & 85.6 & 85.3 & 84.1 & 80.3 & 83.8 \\
FastV~\cite{fastv}                         & \textbf{92.6} & \textbf{89.8} & 96.2 & 83.2 & 90.5 \\
\textbf{GeoScaffold (ours)}                & 90.4 & 86.0 & \textbf{96.8} & \textbf{90.0} & \textbf{90.8} \\
\midrule
\rowcolor{gray!15}
\multicolumn{6}{c}{\textit{keep $64/512$ tokens per image, pruning rates = \textbf{\textcolor{blue}{87.5\%}}}}\\
VLA-Cache~\cite{vla_cache}                 & 57.3 & 58.7 & 64.8 & 26.2 & 51.8 \\
FastV~\cite{fastv}                         & \textbf{91.4} & 66.8 & \textbf{91.8} & 40.4 & 72.6 \\
\textbf{GeoScaffold (ours)}                & 90.0 & \textbf{72.2} & 91.0 & \textbf{66.4} & \textbf{79.9} \\
\bottomrule
\end{tabular*}
\end{table}

Table~\ref{tab:oft_results} reports per-suite success rates on OpenVLA-OFT at two pruning rates. At a moderate pruning rate of $R=0.75$, GeoScaffold reaches a $90.8\%$ four-suite average, slightly ahead of FastV ($90.5\%$) and clearly above VLA-Cache ($83.8\%$). The lead is small in aggregate but concentrates on the spatially demanding suites: GeoScaffold improves over FastV by $6.8$ percentage points on LIBERO-Long and recovers $96.8\%$ on LIBERO-Goal, within $0.5$ pp of the unpruned baseline. As the pruning rate increases to $R=0.875$, the gap widens substantially. GeoScaffold maintains a $79.9\%$ average success rate, while FastV drops to $72.6\%$ and VLA-Cache collapses to $51.8\%$. The contrast is most pronounced on LIBERO-Long, where success rates form a monotonic chain of VLA-Cache $26.2\%$, FastV $40.4\%$, and GeoScaffold $66.4\%$, mirroring the pattern observed on $\pi_{0.5}$. These results suggest that the benefit of preserving spatial scaffolds is not tied to a specific VLA architecture but reflects a general property of robust visual token pruning under aggressive token budgets.

\section{Latency and efficiency analysis}
\label{app:latency}

\paragraph{Theoretical complexity.}
Let $L$ be the prefix length before pruning and $L_r$ the length after pruning at layer $K$, with $D$ the hidden dimension and $M$ the FFN dimension. Each Transformer layer has standard FLOP cost $\mathrm{FLOPs}_\text{layer} \approx 4LD^2 + 2L^2 D + 2LDM$. Pruning to length $L_r$ reduces every subsequent backbone layer's cost by $\Delta\mathrm{FLOPs}_\text{layer} \approx 4(L{-}L_r)D^2 + 2(L^2{-}L_r^2)D + 2(L{-}L_r)DM$, and the action expert benefits at every denoise iteration through its cross-attention to the shortened visual prefix. The pruning overhead itself reuses the layer-$K$ attention map (no extra forward pass) and is dominated by a single \texttt{scatter\_add} and a vectorized FPS lookup, giving total cost $\mathcal{O}(nT)$, several orders of magnitude smaller than one transformer-layer attention $\mathcal{O}(L^2 D)$. Since $L{-}L_r$ scales linearly with the pruning rate $R$, the FLOP reduction grows monotonically with $R$ and translates directly into the prefill speedups measured below.

\paragraph{Empirical latency.}
We profile single-step $\pi_{0.5}$ inference on one RTX 4090, averaging over $100$ LIBERO-Spatial episodes after discarding the initial \texttt{torch.compile} autotune steps. Table~\ref{tab:latency} reports the per-stage latency breakdown for the unpruned baseline and GeoScaffold at three pruning rates.

\begin{table}[H]
\caption{Per-stage single-step latency (ms) of $\pi_{0.5}$ on RTX 4090 under different pruning rates.}
\label{tab:latency}
\centering
\small
\setlength{\tabcolsep}{6pt}
\renewcommand{\arraystretch}{1.10}
\begin{tabular}{lcccccc}
\toprule
Method & image\_encode & prefill & denoise & \textbf{total} & prefill$\uparrow$ & total$\uparrow$ \\
\midrule
Baseline (no pruning) & 11.28 & 27.68 & 22.08 & \textbf{61.03} & $1.00\times$ & $1.00\times$ \\
GeoScaffold $R{=}0.80$  & 11.19 & 15.58 & 20.80 & \textbf{47.56} & $1.78\times$ & $1.28\times$ \\
GeoScaffold $R{=}0.875$ & 11.20 & 14.22 & 20.68 & \textbf{46.10} & $1.95\times$ & $1.32\times$ \\
GeoScaffold $R{=}0.90$  & 11.20 & 14.11 & 20.63 & \textbf{45.92} & $1.96\times$ & $1.33\times$ \\
\bottomrule
\end{tabular}
\end{table}

The latency breakdown in Table~\ref{tab:latency} decomposes a single inference step into three stages: \emph{image encode} corresponds to the SigLIP/Gemma vision encoder, \emph{prefill} to the multimodal backbone forward where pruning takes effect at layer $K{=}2$, and \emph{denoise} to the 10-step flow-matching action expert. The prefill stage benefits most directly from token reduction: GeoScaffold reduces prefill latency from $27.68$ ms at the unpruned baseline to $15.58$ ms at $R{=}0.80$, $14.22$ ms at $R{=}0.875$, and $14.11$ ms at $R{=}0.90$, corresponding to prefill speedups of $1.78\times$, $1.95\times$, and $1.96\times$. The marginal gain saturates above $R{=}0.875$, since the residual prefill cost is dominated by the unprunable image encoder and the text/action prefix. In contrast, denoise latency remains nearly flat across all settings (within $5\%$), because the flow-matching action expert is bounded by attention over the fixed-size action queries rather than by the shortened visual prefix. As a result, the end-to-end speedup plateaus at $\sim\!1.33\times$, set by the prefill share of a single inference step. Overall, GeoScaffold delivers practical acceleration on commodity hardware while keeping the additional pruning overhead negligible relative to a single transformer-layer attention.

\section{Limitations}
\label{sec:limitations}

This work has several limitations. First, our main evaluation is conducted on LIBERO simulation benchmarks, and we have not yet validated the method on real-robot deployments. Second, GeoScaffold assumes that visual tokens retain an underlying two-dimensional patch-grid structure, which may not directly apply to architectures with irregular visual tokenization or learned token merging. Third, although the coverage radius strongly correlates with pruning-induced error in our experiments, it is not the only factor determining VLA performance; semantic relevance, camera viewpoint, manipulation stage, and task difficulty can also affect the final success rate. Finally, our cross-architecture validation covers $\pi_{0.5}$ and OpenVLA-OFT, and broader evaluation on additional VLA families remains future work.

\section{Broader Impact}
\label{sec:broader_impact}

GeoScaffold aims to improve the inference efficiency of vision-language-action models by reducing redundant visual tokens without additional training. A positive impact is that more efficient VLA inference can lower computational cost and make embodied AI research more accessible. At the same time, more efficient robot policies may also accelerate deployment in settings where failures could cause physical harm. Therefore, we do not view inference acceleration as a substitute for safety validation: policies using such pruning methods should still be evaluated under task-specific safety, robustness, and monitoring protocols before real-world deployment.

\section{Existing Assets and Licenses}
\label{app:licenses}

Table~\ref{tab:assets_licenses} summarizes the existing assets used in this work.
We use these assets only for research evaluation and do not redistribute pretrained model weights or benchmark data.
For model checkpoints whose terms depend on upstream base models, we follow the corresponding model-card and repository terms.

\begin{table}[H]
\caption{Existing assets used in this work and their licenses or terms of use.}
\label{tab:assets_licenses}
\centering
\small
\setlength{\tabcolsep}{4pt}
\renewcommand{\arraystretch}{1.10}
\begin{tabular*}{\textwidth}{@{\extracolsep{\fill}}p{0.18\textwidth}p{0.22\textwidth}p{0.50\textwidth}@{}}
\toprule
Asset & License or Terms & Usage in This Work \\
\midrule
$\pi_{0.5}$ / OpenPI
& Apache License 2.0
& Primary VLA model for inference-time pruning evaluation; no redistribution of model weights. \\

OpenVLA-OFT
& MIT License
& Cross-architecture evaluation of the proposed pruning principle; no redistribution of model weights. \\

LIBERO
& MIT License
& Simulation benchmark and evaluation task suites. \\

FastV
& License not specified
& Algorithmic baseline for visual token pruning; no FastV code is redistributed. \\

SparseVLM
& Apache License 2.0
& Visual-token pruning baseline. \\

VLA-Cache
& Apache License 2.0
& Training-free VLA acceleration baseline. \\
\bottomrule
\end{tabular*}
\end{table}

\end{document}